# Towards Interpretable Ensemble Learning for Image-based Malware Detection

Yuzhou Lin[1], Xiaolin Chang[1]

[1] Beijing Key Laboratory of Security and Privacy in Intelligent Transportation, Beijing Jiaotong University, China

**Abstract**—Deep learning (DL) models for image-based malware detection have exhibited their capability in producing high prediction accuracy. But model interpretability is posing challenges to their widespread application in security and safety-critical application domains. This paper aims for designing an *I*nterpretable *E*nsemble learning approach for image-based *M*alware *D*etection (IEMD). We first propose a *S*elective *D*eep *E*nsemble *L*earning-based (SDEL) detector and then design an *E*nsemble *D*eep *T*aylor *D*ecomposition (EDTD) approach, which can give the pixel-level explanation to SDEL detector outputs. Furthermore, we develop formulas for calculating fidelity, robustness and expressiveness on pixel-level heatmaps in order to assess the quality of EDTD explanation. With EDTD explanation, we develop a novel *I*nterpretable *Drop*out approach (IDrop), which establishes IEMD by training SDEL detector. Experiment results exhibit the better explanation of our EDTD than the previous explanation methods for image-based malware detection. Besides, experiment results indicate that IEMD achieves a higher detection accuracy up to 99.87% while exhibiting interpretability with high quality of prediction results. Moreover, experiment results indicate that IEMD interpretability increases with the increasing detection accuracy during the construction of IEMD. This consistency suggests that IDrop can mitigate the tradeoff between model interpretability and detection accuracy.

**Index Terms**—deep Taylor decomposition; dropout; ensemble learning; interpretability; malware detection

## 1 INTRODUCTION

**M**alware binaries, as the critical source of Cyber attacks and threats, have been doing a great deal of harm to Cyber Physic Systems [1]. Deep Learning (DL), as a kind of cyber threat intelligence techniques, has demonstrated its fantastic capability in malware detection [1]. DL-based malware detection approaches can be classified mainly into malicious semantic sequences processing-based and binary visualization textures analysis-based. The former type mainly exploits Natural Language Processing (NLP) techniques to detect malware by using semantic features such as strings, opcodes and application program interfaces (APIs), while the latter type adopts Neuro Network (NN) models to recognize the textures hidden inside malicious images after malware visualization. Leveraging intuitive binary textures learned by NN models is often more effective than extracting highly-processed semantic features after static and dynamic analysis. Binary textures hidden from malware samples might indicate more critical malicious features than those analyzed by NLP methods [4]. This paper focuses on image-based malware detection.

However, NN models are often structurally complex and lack interpretability. Understanding how NNs produce final prediction remains a fundamental challenge [2]. Unless we understand the reason for model decisions, we are unable to apply the image-based approaches widely into security and safety-critical domains [3]. Researchers have devoted efforts to developing interpretable models. However, the previous models [8-14][17-19] were not developed for security-related scenarios, or they [5, 6, 20, 24, 26] were designed on malware datasets but were interpreted in terms of semantic features instead of image features, or they [27] were interpreted in terms of image features but by manually plotting interpretable images to achieve model-agnostic and local explanation results. Moreover, no previous works on image-based malware detectors like [4,16,22,27,39] explored how to use the interpretability results for improving detection accuracy. Note that both local interpretability/explanation techniques and sensitive analysis-based interpretability techniques [8-14] may miss critical hidden malicious behaviors. Layer-wise relevance propagation (LRP) [8,17,30] uses global interpretability technique to reduce these limitations but there is a lack of LRP implementation on malware detection. How to make global explanation on image-based malware detector is critical and requires further exploration.

How to quantitatively evaluate an interpretability technique is also critical. Metrics, including Fidelity [5], Robustness [6] and Stability [6], have been defined. Fidelity aims to indicate the correctness and effectiveness of one explanation. Robustness aims to represent the difference degree of one family explanation from another family explanation. Stability aims to indicate the similarity between the explanations of a set of same pre-trained models under the same interpretation method. It is a fundamental metric to assess the quality of local explanation but is not required for evaluating global explanation because global explanation methods are model-specific to a single model. Their calculation formulas [6] have been defined to assess the quality of interpretability. But these formulas are only applied for the semantic-level interpretability of models, which focuses on sequences of malicious behaviors. It is unsuitable for image-level interpretability, which focuses on pixels and pixel matrices. Note that this paper uses explanation and interpretation interchangeably.

Ensemble learning is a technique of combining several single models to improve a single model's detection accuracy. Still, it has not been widely used in the field of image-based malware detection. The recent work like [4] only combined ResNet50 and VGG16 for fine-tuning operation, while more effective Convolutional NN (CNN) models like GoogLeNet Inception v3 can be utilized for enhancing image-based malware detection results. In addition, there are no interpretability techniques for ensemble learning.

These discussions motivate the work presented in this paper, namely, developing an ensemble learning based detector with high accuracy and high interpretability to suit a security-related case. There are four major challenges in developing an interpretable ensemble learning framework as follows:

**Challenge 1**: How to select an effective ensemble learning group on CNN candidates for a high malware detection rate? An inappropriate grouping method either leads to large time cost (for example, exhaustive ensemble strategy) and/or is unable to improve accuracy.

**Challenge 2**: How to effectively interpret an ensemble learning model with high-quality explanation results?

**Challenge 3**: How to quantitatively assess the quality of our image-based explanation method from its fidelity [5], robustness [6], expressive power [3]?

**Challenge 4**: How to mitigate the tradeoff between malware detection accuracy and detector interpretability? A more interpretable model often performs less accurate detection such as using decision trees to classify families [3].

The following summarizes our main contributions, which tackle the above challenges.

- We design a *Selective Deep Ensemble Learning*-based (SDEL) detector, including two-level classifiers. By selective and deep, we mean that only CNN models with different deep NN structures are selected to establish the first-tier classifiers. This difference of CNN structures is measured by a novel distance computation formula, defined in Eq. (2) of Section 3. The second-tier classifier uses the stacking technique [36] to extract and combine the features from the first-tier classifiers and then fine-tune a fully-connected layer with random dropout.
- We propose an *Ensemble Deep Taylor Decomposition* (EDTD) approach, for globally and reliably interpreting SDEL detector to help understand the detector prediction decision. To the best of our knowledge, we are the first to customize *Deep Taylor Decomposition* (DTD) algorithm on an ensemble learning model under a case study of interpreting malware detection.
- We propose a metric to evaluate expressive power [3], namely, Expressiveness, and also define the formulas of calculating Fidelity [5], Robustness [6] and Expressiveness to assess the quality of SDEL detector interpretability. It is the first time to define these formulas for ensemble learning pixel-level interpretability.
- We propose an *Interpretability*-based *Drop*out (IDrop) approach, which uses EDTD explanation to train SDEL detector so as to develop IEMD. IDrop effectively enhances detection results of SDEL detector from 97.16% to 99.87%. Meanwhile, IEMD exhibits interpretability with high quality of prediction results. Moreover, IDrop can mitigate the tradeoff between model interpretability and detection accuracy. This is validated by experiment results of **Fig.9**, in which IEMD interpretability is improved with the increasing detection accuracy during the construction of IEMD.

The rest of the paper is organized as follows: Section 2 surveys the related work of malware visualization detection, interpretable approaches in image recognition, and dropout techniques. Section 3 describes the main workflow of achieving IEMD. Section 4 illustrates the conducted experimental results and Section 5 draws a conclusion and discusses future trend of interpretable malware detection.

## 2 RELATED WORK

This section first presents previous ML models for image-based malware detection and then interpretability techniques related to our work are discussed. **Table 1** presents the comparison between these previous researches and our work.

### 2.1 Image-based ML Models for Malware Detection

Model-based malware detections can be semantic feature-based or image-based. Semantic feature-based methods apply features like API/Strings/Opcodes/permissions/memory for classification but image-based methods usually perform better without much expert knowledge and the complicated process of extracting malware features. Image-based methods can be categorized into traditional ML-based and DL-based methods. For the former type, GIST and *Gray-Level Co-occurrence Matrix* (GLCM) features were first extracted from grayscale images and then fed up into models like SVM/KNN [22, 23]. The latter type like [8] fed up grayscale images converted by the original binaries directly into DL models like CNN architectures. The former type of methods can achieve acceptable precision results but the latter type performs much better. All these works only considered a single CNN and their interpretability is not provided.

Recently, ensemble learning technique is explored to develop imaged-based malware detectors like [4], which can effectively detect the packed and unpacked malware but lack the model interpretability. Our proposed IEMD based on ensemble learning not only outperforms their detectors of [4] but also is interpretable.

TABLE 1 COMPARISON ON THE WORK DISCUSSION IN SECTION 2

| Ref. | Dataset type | | | Classifier | Ensemble learning | Explanator | Interpretability assisting detection |
|---|---|---|---|---|---|---|---|
| | Image | Image-based malware | Semantic feature-based malware | | | | |
| [4] | √ | √ | × | DL | √ | × | × |
| [22, 23] | √ | √ | × | ML | × | × | × |
| [8-14] | √ | × | × | DL | × | Sensitivity analysis | × |
| [8, 17] | √ | × | × | DL | × | DTD | × |
| [19] | √ | × | × | DL | × | LIME | × |
| [28] | √ | × | × | DL | × | SHAP | × |
| [5] | × | × | √ | DL | × | LEMNA | × |
| [6] | × | × | √ | DL | × | SP-LIME | × |
| [18] | × | × | √ | DL | × | LIME | × |
| [20] | × | × | √ | DL | × | XMAL | × |
| [24] | × | × | √ | DL | × | I-MAD | √ |
| [25] | × | × | √ | DL | × | IF-THEN rules | × |
| [26] | × | × | √ | ML | × | Random forest | √ |
| [27] | √ | √ | × | DL | × | PLOTsuper-pixels | × |
| [21] | × | × | × | DL | × | DTD | √ |
| **IEMD** | √ | √ | × | **DL** | √ | **EDTD** | √ |

- 'Image' denotes that the dataset studied is image-based.
- 'Image-based malware' denotes that malware dataset is image-based.
- 'Semantic features-based malware' denotes that dataset is for malware but not visualized, and semantic features are used for prediction.
- 'Classifiers' denotes that the detection classifier used for malware predictions is traditional machine learning-based (ML) or deep learning-based (DL).
- 'Ensemble learning' denotes whether the ensemble learning technique is applied.
- 'Explanator' denotes whether an explanation method is used and what kind of interpretability techniques is used.
- 'Interpretability assisting detection' denotes whether an explanation method is used to enhance detection performance.

TABLE 2 COMPARISON ON INTERPRETABILITY TECHNIQUES FOR IMAGE-BASED MODELS

| Explanator Type | Explanator | Scale | Model dependence | Image Mal | Quantified Fidelity | Ensemble leaning |
|---|---|---|---|---|---|---|
| SA | $\nabla o(x)$ [8] | L | A | × | × | × |
| | $(\nabla o(x))^2$ [8] | L | A | × | × | × |
| | $\nabla o(x) \odot x$ [8] | L | A | × | × | × |
| | smooth $\nabla o(x)$ [11] | L | A | × | × | × |
| | Occlusion analysis [14] | L | A | × | × | × |
| | Vanilla $\nabla o(x)$ [9] | L | S | × | × | × |
| | guided $\nabla o(x)$ [10] | L | S | × | × | × |
| | guided smooth $\nabla o(x)$ [12] | L | S | × | × | × |
| | grad CAM [13] | L | S | × | × | × |
| | guided grad CAM[13] | L | S | × | × | × |
| LIME | LIME [19] | L | A | × | × | × |
| | PLOTsuper-pixels [27] | L | A | √ | × | × |
| SHAP | kernel SHAP [29] | L | A | × | × | × |
| | tree SHAP [28] | L | S | × | × | × |
| LRP | DTD [8, 17] | G | S | × | × | × |
| | DeepLIFT [30] | G | S | × | √ | × |
| | EDTD (Our work) | G | S | √ | √ | √ |

- L and G, respectively, denote local and global interpretability.
- A and S, respectively, denote model-agnostic and model-specific.
- 'Image Mal' denotes whether this explanator is for image-based malware dataset.

## 2.2 Model Interpretability Work towards Image Recognition

An interpretability technique aims to find out the effect of model input features on output decisions, giving the reason why there is a relationship between input and output [3]. Malware's semantic features like assembly code and blocks have been used to interpret malware-detection models [5, 6, 18, 20, 24-26]. This section focuses on model interpretability techniques towards image datasets, which, according to their working principles, can be classified into sensitivity analysis (SA) techniques, local interpretable model-agnostic explanation (LIME), SHAP and Layer-wise relevance propagation (LRP).

**Table 2** summarizes their comparison. 'Model-dependence' represents whether a technique is model-specific or model-agnostic type. For the former type, interpretation only suits a specific model such as linear regression models. For the latter type, interpretation can be applied on any model without considering the model's inner structures. There are two interpretation scales: global and local. Global interpretation requires opening black-boxes and then obtaining both algorithms and weights. Local interpretation only helps understand a part of instances or a single data point without even giving insight into the model architecture. Global interpretation becomes more reliable than local interpretation [3]. Below we shortly summarize the major principles behind these explanators.

SA-based interpretability techniques adopt human observation to obtain the contribution of input features to output by model sensitivity based on gradients or attention mechanisms. Then the interpretability results are uncertain and local. Specifically, $\nabla o(x)$ [8] calculated the input/output-layer gradients. $(\nabla o(x))^2$ [8] calculated the square of the input/output-layer gradients. $\nabla o(x) \odot x$ [8] multiplied the input/output-layer gradients with input features. Smooth $\nabla o(x)$ [11] added noise to input images to conversely optimize gradient results. Grad CAM [13] used both each layer gradient and weights to improve class activation mapping (CAM) (note that CAM must alter inner architectures of one specific CNN). Occlusion analysis [14] used sliding occlusion windows to enhance sensitivity analysis. Vanilla $\nabla o(x)$ [9] backpropagated through layers of a specific model to get gradients of the input layer. In addition, guided-type SA exploits guided backpropagation [10], which slightly differs from conventional backpropagation because guided backpropagation visualizes gradients with respect to the image where negative gradients are suppressed through ReLU layers. Followed by the above depiction, guided smooth $\nabla o(x)$ [12] added noise on the input data and then used the same mechanism as that in [10]. Guided grad CAM [13] used the guided backpropagation technique to further improve grad CAM.

LIME [19] achieved local interpretation by finding out local points of input features, and using data argumentation to fit a new intrinsically interpretable model as surrogate, which is interpretable to the selected partial feature space. Chen *et al.* [27] used LIME to plot super-pixels of malware images. SHAP [28, 29] explored cooperative game theory to get average of the contributions (denoted by Shapley Value) of all features and used the average to get local interpretability results.

LRP techniques [8, 17, 30] decomposed output relevance backwards in the neural network. More precisely, output relevance equals to output vectors, which are backward propagated in the output layer, intermediate layers, input layers and finally to the input images, and then are considered as the contribution scores of different input features. The details of LRP schematic equations are depicted in [17].

Extensions to the LRP technique were proposed. Shrikumar *et al.* [30] calculated the difference between reference activation and real activation so as to backpropagate this difference (denoted as relevance) to the input features in a layer-wise manner, denoted as DeepLIFT. The authors in [8] and [17] used LRP with deep Taylor decomposition (DTD) to obtain input feature contribution. Unlike DeepLIFT, DTD–based explanation techniques aim to calculate relevance by the following process.

Let $y_k$ be the relevance of neuron $k$ at one layer and $\Lambda = (h_j)_j$ denote neuron activations in the lower layers. DTD computes $y_k$ by using a function $f_k$ of $\Lambda$ and trying to find out the root point $\widetilde{\Lambda}$ through Taylor expansion at this root point as conducted by Eq. (1).

$$f_k(\Lambda) = f_k(\widetilde{\Lambda}) + \sum_j [\nabla f_k(\widetilde{\Lambda})]_j \cdot (h_j - \widetilde{h}_j) + \cdots \quad (1)$$

The linear term $[\nabla f_k(\widetilde{\Lambda})]_j \cdot (h_j - \widetilde{h}_j)$ defines the share of $y_k$ redistributed to neuron $j$ in the lower layer. In order to solve $f_k(\Lambda)$, $f_k$ is replaced by a "relevance model" $\hat{f}_k$ that satisfies $\hat{f}_k(\Lambda) \approx f_k(\Lambda)$ with local information at the current activations $\Lambda$. Note that $\widetilde{\Lambda}$ satisfying $f_k(\widetilde{\Lambda}) = 0$ refers to the root point which is required to find out. More details can be found in [17].

The previous DTD-based interpretability techniques like [8, 17], interpreted a single neural network from the output layer to the input layer. They cannot be used for ensemble learning-based detectors. Ensemble learning models exploit feature incorporated by different base classifiers. The conventional DTD techniques expand and decompose relevance in a layer-wise manner without parallel redistribution and backpropagation, which suggests its unsuitableness on ensemble architecture. However, our EDTD approach exploits not only relevance backpropagation mechanism but also relevance split and combination to explain the SDEL detector prediction.

### 2.3 Dropout to Enhance Model Performance

The conventional dropout (namely, random dropout) [31] used randomly setting to drop out neurons so as to train different neural networks in each epoch. The previous experiment results of a wide range of datasets demonstrated its capability in preventing overfitting and enhancing model performance.

Comparatively, Wan *et al.* [32] dropped out some weights rather than input neurons to outperform the conventional dropout method. Furthermore, Wang *et al.* [33] introduced a defensive dropout mechanism against adversarial attacks to enhance model robustness. The authors in [15, 34, 35] proposed novel extended dropout mechanisms, which are more adaptive to certain models. All these dropout methods did not leverage model interpretability to enhance training results.

Recently, Schreckenberger *et al.* [21] explored using relevance score calculated by DTD to improve the conventional dropout technique, which can enhance model robustness and prevent overfitting better than using random dropout. But it did not use image datasets. Since heatmaps [37] generated by DTD provide reliable explanations and the related explanation results have been quantitatively evaluated by our experiment comprehensively, we hereby propose interpretable dropout (IDrop) approach, which can enhance detector performance while exhibiting reliable pixel-level explanations.

## 3 METHODOLOGY OF ACHIEVING IEMD

An overview of achieving IEMD is roughly shown in **Fig.1**. Firstly, SDEL approach generates SDEL detector. Then EDTD approach interprets the prediction results of SDEL detector. Lastly, we exploit IDrop approach to further optimize explanations on SDEL detector and further enhance detector performance. In the following, the entire workflow is detailed.

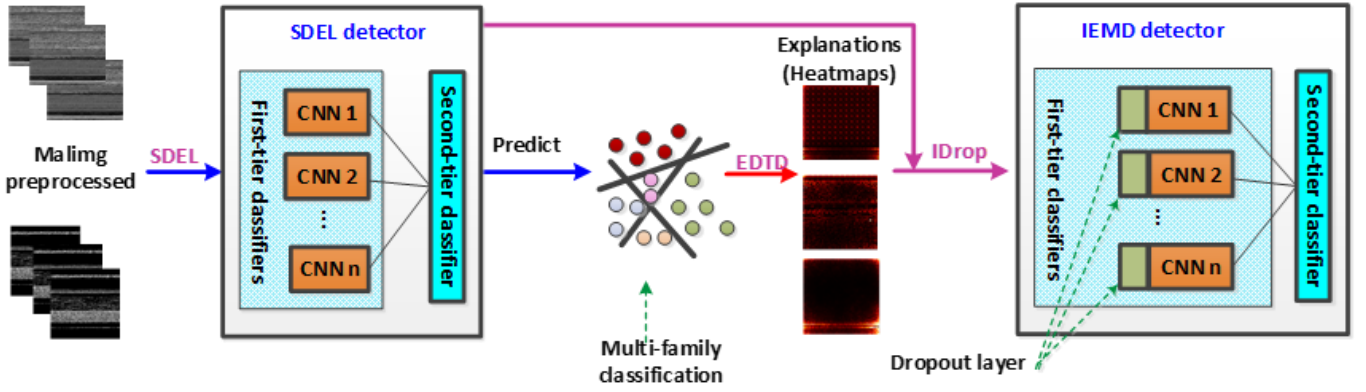

Fig.1 An overview of achieving IEMD.

## 3.1 SDEL Detector

### 3.1.1 Dataset and preprocessing

We consider the malware dataset called Malimg [23], which contains 9339 malware grayscale images without binary raw files from 25 families. All images are resized into the size of 256 ×256. Specifically, the input images are resized to 299×299 so as to satisfy the requirements of GoogLeNet Inception v3. The whole dataset is divided into two parts, 80% (7470 samples) to the training set and the left 20% (1869 samples) to the test set. Note that Malimg samples are imbalanced as shown in [23]. Features are similar in a family but there are significant differences in different families. To overcome data imbalance, we use the stratified validation strategy [43] (note that we randomly select samples with the same scale number in each family) to randomly combine the training set with all families. On the training set, we give the 10-fold stratified cross validation to prevent the overfitting and imbalance of data distribution. On the test set, we set the same data distribution with that of training set.

### 3.1.2 SDEL approach

**Fig.1** indicates that SDEL detector comprises of two-level classifiers, the first-level and the second-level. There are $n$ first-level base classifiers and only one second-level classifier. **Algorithm 1** describes SDEL approach. First, we customize 8 baseline CNN models pre-trained on ImageNet by adding a dropout layer and a fully-connected (FC) layer containing 25 classes from Malimg instead of the final FC layer (intended for 1000 classes from ImageNet). We then fine-tune the input layer, convolution layers and FC layers of all baselines (by line 4 of **Algorithm 1**). After fine-tuning the base classifiers for selection, we remove the last linear layers of all classifiers so as to extract the feature vectors generated from first-level classifiers, and add a new FC layer with random dropout to be trained (by line 5 of **Algorithm 1**). This new FC layer is the second-level classifier. The random dropout function makes the last fully-connected layer a set of ensemble classifiers with different neurons to learn the feature space. Note that similar models are not considered because similar structures are hard to enhance ensemble learning results a lot according to the basic strategy proposed in [36]. Furthermore, in line 10 of **Algorithm 1**, we define $distance(c_i, c_j)$, where $c_i$ $c_j$ are respectively from base first-level classifiers being fine-tuned. This function would measure the distance of two classifiers confusion matrix on test set. For instance, if $c_i$ and $c_j$ detect very different families correctly, then their distance is very high but not higher than 1 (range is from 0 to 1). The detail calculation is by Eq. (2):

$$distance(c_i, c_j) = \frac{\sum_{k=0}^{24}(||r_i(k)|-|r_j(k)||)}{\sum_{k=0}^{24}(|r_i(k)|+|w_i(k)|)} \qquad (2)$$

Where, $r_i(k)$ denotes the correct detection results for family $k$ by $c_i$ and $|r_i(k)|$ denotes the sample number of $r_j(k)$. $w_i(k)$ denotes the incorrect detection results likewise. Then we select top3 distance results of different CNN structure models to be ensemble (by line 11, 16 in **Algorithm 1**), and select the best accuracy group (by line 19-25 in **Algorithm 1**).

---

**Algorithm 1 SDEL Approach**

Input: $C = \{c_1, c_2, \ldots, c_n\}$: pretrained base CNN classifiers as first-level modules; $M$: second-level classifier;
Output: selective ensemble learning module of IEMD: $E$;
1. initial ensemble learning selection stack: $ES \leftarrow \emptyset$; ensemble learning base number: $k \leftarrow n$; initial final selection output model: $EM \leftarrow \emptyset$;
2. while not empty in $C$ do
3.     pop up $c_i$ and do
4.     transfer train ($c_i$); note that fine-tune the input layers, convolution, FC layers;
5.     extract feature space and add dropout ($c_i$);
6. end while
7. for $k$=0 to $n$ do
8.     for $i$=0 to $n$ do
9.        for $j = i + 1$ to $n$ do
10.           if $c_i$ is not similar to $c_j$ and $distance(c_i, c_j)$ is in top3 then
11.             $ES \leftarrow \{c_i, c_j\}$;
12.           end if
13.        end for
14.     end for
15.     if $length\ (ES) = k$ then

16.         EM←fine-tune (ES, M); note that fine-tune the linear layers
17.     end if
18. end for
19. while not empty in **EM** do
20.     pop up $em_i$ in **EM** and do
21.     if $em_i$ shows highest accuracy then
22.         $E \leftarrow em_i$;
23.     end if
24. end while
25. return **E**.

## 3.2 Interpretation Approach to SDEL Detector

There are three major challenges in designing a DTD-based interpretability approach to SDEL detector as follows:
- First, detector output relevance should be redistributed properly to the first-tier base classifier output layers, which requires relevance splitting and redistribution.
- Next, different CNN classifiers may have different inner structures. The expansion and decomposition on the layer-wise relevance through the first-tier classifiers should consider these structure differences.
- Finally, how to combine the input final relevance and consider the contribution of different first-tier CNN models to ensemble learning outputs.

This section first presents how EDTD tackles these challenges and then interpretation evaluation metrics are presented.

### 3.2.1 EDTD approach

**Algorithm 2** describes EDTD approach, which interprets SDEL's detection results and gives a pixel-level explanation represented in the form of saliency heatmaps [37]. The ensemble neural network prediction from malware grayscale images is then subject to EDTD which propagates the prediction results backward in the network like LRP, and the outcome is converted to heatmap where brighter pixels indicate the more suspicious part in binary images. Note that $w_{ij}$ and $bi$ are the weights and bias in the $i$th layer where the former layer is the $j$th layer. $X_f$ is the prediction result, and $R_f$ is the relevance score of prediction. $R_i$ is the $i$ th layer Taylor expansion root point and decomposed to the next layer as $R_{j1}$ to $R_{jn}$.

---

**Algorithm 2 EDTD Approach**
**Input:** $p$: detection results of SDEL; $M$: SDEL model; $h$: heatmap scale;
**Output**: saliency heatmap $H$;
1. initial relevance score: $R \leftarrow \emptyset$ and flatten module in $M$ to stack: $S \leftarrow \{m_1, m_2, \dots, m_n\}$;
2. while not empty in $S$ do
3.     pop up $m_i$ and do
4.     if length ($S$) = 0 then
5.         if $m_i$ is linear then
6.             $R \leftarrow$ calculate relevance ($m_i$, $R$, $p$);
7.         else if $m_i$ is convolution layer then
8.             $R \leftarrow$ calculate relevance ($m_i$, $R$);
9.         else if length ($S$) ≠ 0 then
10.         if $m_i$ is in base classifier $C_i$ then
11.             $R \leftarrow$ calculate relevance ($m_i$, $R$, $C_i$);
12.         else if $m_i$ is adaptive average pool then
13.             $R \leftarrow$ calculate relevance ($m_i$, $R$);
14.         else $m_i$ is linear and $m_i \in$ base classifier $C_i$ then
15.             $R_i \leftarrow$ calculate relevance for entrance of $C_i$;
16.     end if
17. end while
18. for $i$=0 to $n$ do
19.     $R \leftarrow$ combinate relevance($R_i$);
20. end for
21. generate saliency heatmap $H$: $max$ (0, $R$)∗255∗ $h$;
22. return $H$.

---

We now explain **Algorithm 2** with the definition and formulas defined in Section 2.2.3. Let $v_{ij} \rightarrow a_i - \tilde{a}_j$ be the root search direction. Then, we obtain a closed-form solution with $R_{i \leftarrow j} = v_{ij}w_{ij}R_j/(\sum_i v_{ij}w_{ij})$, where we can decompose relevance from the $j$-th neuron to the $i$-th neuron by expansion and decomposition through Eq. (1) in Section 2.2.3 (note that $w_{ij}$ refers to weights from $i$th to $j$th layer). Each pixel with a propagated relevance is reliable because at each layer we find out the root point that satisfies $R_j\left((\tilde{a}_i)_i^{(j)}\right)$=0. In the second-level classifier, we first propagate the relevance through the FC layer and split the relevance into $n$ parts, each of which is backpropagated to the corresponding first-level classifier. This is achieved by the line of 15 in **Algorithm 2**.

Concretely, although each different CNN model such as ResNet, VGG or SqueezeNet can have different layers and units. All these units and blocks consist of basic layers such as normal layer, dropout layer, relu layer, maxpooling layer and etc., which all can be calculated in the line 5-16 from **Algorithm 2**. The details of backpropagation are in [17].

Finally, let output neuron $i$ in model $k$ have a corresponding input relevance score $R(i,k)$. Let $m,n$ respectively denote the number of input dense layer neurons in one model and the total number of first-level classifiers, and $p_k$ represent the detection accuracy given by a model $k$. Then we propose a novel weighted averaging calculation strategy to generate relevance score $\hat{R}$ of all first-level classifier models, calculated in Eq. (3):
$$\hat{R} = \sum_{\substack{0 \leq i \leq m \\ 0 \leq k < n}} (R(i,k) \cdot p_k/(\sum_{j=0}^{n} p_j)$$
(3)

As is depicted above, $p_k/(\sum_{j=0}^{n} p_j)$ attributed to base first-level model $k$ decides the final redistribution and combination of EDTD explanation results (conducted by line (19) from **Algorithm 2**). Hereby, we calculate relevance score from the output backward to the input and then obtain a vector consisting of the sum of different base classifiers relevance scores in ensemble learning. The achieved vector with the size of input images indicates the relevance of each input based on the data.

Now we can identify the locations of interest generated by the EDTD interpretation, and refer back to the binary file to identify new patterns and malicious signatures. Because our dataset is Malimg, which merely consists of grayscale images, we give pixel-level explanation strategy for future use on other datasets. Besides, unlike attention mechanism [20] and super pixels plotted by LIME [27] used for malware detection interpretability, our EDTD algorithm gives global insights into opening ensemble black-boxes and also comprehensively validated by quantitative indicators

### 3.2.2 Interpretability evaluation metrics

We assume that there are $k$ malicious pixels and $m$ is defined to denote an explanator. In Section 4.3, we investigate 14 explanators listed in **Fig.4**. For explanator $m$, each malicious pixel $i$ has a relevance $f_i(m)$. $\tau$ denotes a pre-defined threshold of lower relevance. Then, we obtain the set $\Omega(m,\tau)$ of the malicious pixels to be flipped in Eq. (4).

$$\Omega(m,\tau) = \{f_i(m)|f_i(m) \geq \tau\} \quad (4)$$

With these definitions, we now present the formulas for calculating Fidelity, Robustness and Expressiveness on heatmaps.

**Fidelity**. First, we calculate average detection accuracy $a_m(\tau,i)$ with flipping a set of pixels on $\Omega(m,\tau)$ by $i$ ($i \leq n$) times, and each detection accuracy of pixel-flipping procedure on $z$ from the test set $\mathbf{Z}$ at the $i$-th time is denoted as $pf(\Omega(m,\tau),i,z)$ in Eq. (5).

$$a_m(\tau,i) = (\sum_{\forall z \in \mathbf{Z}} pf(\Omega(m,\tau),i,z))/|\mathbf{Z}| \quad (5)$$

For $\forall z \in \mathbf{Z}$, if the prediction is correct, then $pf(\Omega(m,\tau),i,z)$=1, otherwise $pf(\Omega(m,\tau),i,z)$=0. With the pixel-flipping curves of $a_m(\tau,i)$, we calculate Fidelity $FD_m(\tau)$ from curves, which is generated by finding and summing the partial derivatives of $i$ as shown in Eq. (6). Multiplying '100' aims to enlarge the fidelity score for better observation and '$n$' aims to obtain an average number of fidelity score by $n$ times.

$$FD_m(\tau) = 100 * (\sum_{i=1}^{n}||\frac{\partial a_m}{\partial i}||)/n \quad (6)$$

**Robustness**. We first define the similarity between different samples used for explanation. $e_i^{|\Omega(m,\tau)|}(m)$ denotes the $i$-th sample heatmap generated by explanator $m$. Similar definition is for $e_j^{|\Omega(m,\tau)|}(m)$. The similarity is obtained by Eq. (7).

$$sim\left(e_i^{|\Omega(m,\tau)|}(m), e_j^{|\Omega(m,\tau)|}(m)\right) = \quad (7)$$
$$\begin{cases} \sum_r(|Rel(e_i^{|\Omega(m,\tau)|}(m),r) - Rel(e_j^{|\Omega(m,\tau)|}(m),r)|) \\ \qquad , if\ r \in e_i^{|\Omega(m,\tau)|}(m) \cap e_j^{|\Omega(m,\tau)|}(m) \\ \sum_r|Rel(e_i^{|\Omega(m,\tau)|}(m),r)| \ , if\ r \notin e_i^{|\Omega(m,\tau)|}(m) \cap e_j^{|\Omega(m,\tau)|}(m) \end{cases}$$

where $Rel(e_i^{|\Omega(m,\tau)|}(m),r)$ denotes the relevance score of pixel $r$ in the heatmap generated by explanator $m$ for sample $i$. 'sample' denotes the individual instance of one malicious family. Note that we should consider whether pixel $r$ is in the intersection set of two heatmaps used for calculation. If pixel r is not in the intersection set, then we treat the relevance of pixel r in the $i$-th sample as the result of subtraction between the $i$-th and $j$-th samples. Then we obtain the different heatmaps similarity.

High robustness indicates that different families should have very different explanation heatmaps for excellent intuitive representation. Let $\Phi$ be the set of data of the same predicted family, and $\Psi$ be the set of data from different predicted families. $\Phi$ and $\Psi$ are respectively obtained by the original test set. $\eta(\Phi)$ and $\eta(\Psi)$ respectively denote the number of times of iteratively calculating similarity on each set. Then we calculate robustness by using Eq. (8), in which the smaller the value, the better the robustness.

$$RB_m(\tau) = \frac{\sum_{x_{i,j} \in \Phi} sim\left(e_i^{|\Omega(m,\tau)|}(m), e_j^{|\Omega(m,\tau)|}(m)\right)/\eta(\Phi)}{\sum_{x_{i,j} \in \Psi} sim\left(e_i^{|\Omega(m,\tau)|}(m), e_j^{|\Omega(m,\tau)|}(m)\right)/\eta(\Psi)} \quad (8)$$

**Expressiveness**: An effective image-based explanator should turn light on expressive power [3] for human to correctly and easily to distinguish heatmaps. This indicator is redefined as expressiveness, which can be used to express the correctness and discrimination of interpretation methods, because more discriminative results help human better distinguish and correct results make this help effective. In other words, expressiveness is a proper combination of fidelity and robustness. We propose to use Expressiveness to evaluate expressive power. $EP_m(\tau)$ is defined to denote Expressiveness of explanator $m$ under the chosen relevance threshold $\tau$. Eq. (9) gives the calculation formula.

$$EP_m(\tau) = FD_m(\tau)/RB_m(\tau) \quad (9)$$

Eq. (9) indicates that the better the fidelity or robustness, the larger the expressiveness score, namely, the better the expressive power.

### 3.3 IDrop Approach for Generating IEMD

From EDTD approach, we can obtain heatmaps, whose quality is demonstrated by our Section 4.3 experiment results in terms of Fidelity, Robustness and Expressiveness. In this section, we present IDrop approach which generates IEMD by using these heatmaps to enhance the detection capability of SDEL detector. **M** denotes SDEL model and $\mathbf{I} = \{i_1, i_2, ..., i_n\}$ denote the set of preprocessed grayscale images. **Algorithm 3** describes IDrop approach.

We first present how to generate **M** and its explanation **HS**, described in line 2-22. We use the 10-fold stratified cross-validation and set the number of epochs with 40 in each fold. In addition, EDTD recalculation is conducted every 5 epochs and pushed into a stack **RS** (by line 6) and this RS will be cleared every 5 epochs by line 4. According to the relevance heatmaps calculated by EDTD, we calculate the dropout mask **MASK**, which helps drop out the less relevant nodes (by line 7-14). Each input image has its own relevance heatmap with the same shape of image scale (no matter whether its prediction is correct or not). Next, we generate a set of random variables with uniform distribution between zero and the maximum value of relevance score. Given the distribution having the same shape with that of images, we check whether the relevance is greater than the randomly-drawn value. If it is the case, the corresponding neuron is kept for training. Otherwise, neurons will be dropped out. Then, we achieve dropout masks towards all samples to further fine-tune input and output layer of each first-tier classifier, and the entire second-tier classifier of SDEL detector (by line 15-18). By line 19, we validate the fine-tuned SDEL detector in this epoch. Then we update the weights which can help obtain the best validation accuracy in this fold.

---

**Algorithm 3 IDrop approach**

Input: **M**, **I**
Output: **IEMD**, **HS**

1.  $U \leftarrow \emptyset$, **MASK** $\leftarrow \{m_1 = 0, m_2 = 0, ..., m_n = 0\}$, h ← 5000
2.  for each fold in all do
3.    for each epoch in each fold do
4.      if epoch mod 5=0 then **RS**← ∅
5.        for each image $i_j$ in **I** do
6.          **R** ← EDTD ($i_j$, **M**, $h$) and push **R** into **RS**
7.          **U** ← $uniform\ distribution\{0, max(R)\}$
8.          for $u$ in **U** and $r$ in **R** do
9.            if $u < r$
10.              $m_i$ =0
11.            else`
12.              $m_i$ =1
13.          end for
14.        end for
15.        for each $i_j$ in **I** do
16.          $i_j \leftarrow i_j \cap $ **MASK**
17.        end for
18.        **M** ← fine-tune (**I**, **M**) in this epoch;
19.        validate **M** in this epoch and update the best parameters
20.    end for
21.  end for
22.  **HS** ← **RS**
23.  for each $\beta$ in {0,0.1,0.2,0.3,0.4, 0.5, 0.6, 0.7, 0.8, 0.9} do
24.    if $\beta$ improves detection accuracy of **M**
25.      $\beta \leftarrow \beta$
26.  end for
27.  generate **IEMD**: **IEMD** ← {**M**, $\beta$}
28.  return **IEMD**, **HS**

---

The previous 10-fold cross validation actually trains a set of different models with different input neurons to be discarded [31] and **M** is determined from these models after line 21. In line 22, we obtain **HS**, holding the final explanations. This combination of model construction and explanation in line 2-22 to generate the best M help IDrop approach mitigate the tradeoff between detection accuracy and interpretability. We now tune a hyperparameter $\beta$ to make the model **M** without EDTD-based dropout equal to obtain the best performance. This hyperparameter originally denotes the probability of discarding neurons and is validated to be set with 0.5 to obtain the best detection accuracy in [31]. However, in our work, since the dropout mechanism differs from the conventional dropout, $\beta$=0.5 may no longer suit our case. Thus, by line 23-26, we tune it in the detection phase with a value from 0 to 1, and obtain a proper value to help achieve the best detection accuracy. Note that we tune it by multiplying (1-$\beta$) on input layer weights to rescale these weights so as to make the input layer equal to this layer with a dropout layer previously. In line 27, we obtain the final IEMD with {**M**, $\beta$}.

## 4 EXPERIMENTS

We use the Pytorch library to implement several experiments in python3. The hardware environment is set with one NVIDIA T4 GPU card with 8GB, one Intel Xeon CPU with 16 cores and 64 GB main memory for test by a batch-size of 128.

### 4.1 Evaluation Measures

Before defining detection performance metrics, some definitions are given first. Let True Positive (*TP*) denote correct detection of one malware family, True Negative (*TN*) denote correct detection of not being this malware family, False Positive (*FP*) denote false identification of malware family and False Negative (*FN*) denote false identification of not being one malware family. With these definitions, the metrics used to compare the performance of classifiers include accuracy, precision, recall, and F1-score, which are defined as follows.

**Accuracy**. Denote the ratio of correctly predicted outcomes to the sum of all predictions and can be defined as follows:
$Accuracy = (TP + TN)/(TP + TN + FP + FN)$

More specifically, training and validation accuracy are respectively calculated in the phase of *k*-fold cross validation on the training set and validation set, while detection accuracy is calculated on the test set. To evaluate the model performance, detection accuracy matters more.

**Precision**. Determine if the positive predictions of one model are correct and can be calculated by $Precision = TP/(TP + FP)$.

**Recall**. Denote positives identified by the model from all possible positives and is defined by $Recall = TP/(TP + FN)$.

**F1-score**. Denote the harmonic average of recall and precision, which can be defined as:
$F1 = 2 \times (Precision * Recall)/(Precision + Recall)$

In terms of model interpretability, we use metrics proposed in Section 3.2.2 to quantitatively assess the quality of explanation results generated by EDTD, and other baseline explanators, including fidelity, robustness and expressiveness.

### 4.2 Evaluation of SDEL Approach

According to SDEL approach, base classifiers are first fine-tuned on Malimg so as to obtain the selection of first-tier base classifiers. We select eight major popular CNN architecture models, namely, AlexNet, ResNet18, ResNet 50, VGG11_bn, VGG16 [41], DenseNet121, Squeezenet [40] and GoogLeNet inception v3. They are pre-trained on the big dataset called ImageNet (1000 families) and denoted by $c_1, c_2, ...$ and $c_n$, respectively. *n*=8. This section evaluates the effectiveness of SDEL approach by comparing detection accuracy of SDEL detector with that of these 8 base CNN models.

TABLE 3 COMPARISON OF SDEL DETECTOR AND 8 BASELINES TRANSFER TRAINED ON OUR CHOSEN DATASET TO PROCESS 25 FAMILY MULTI-CLASSIFICATION. NOTE THAT TL MEANS TRANSFER LEARNING

| Model | Detection accuracy |
|---|---|
| TL+ResNet18 | 92.08% |
| TL+ResNet50 | 91.63% |
| TL+VGG16 | 90.27% |
| TL+GoogLeNet Inception V3 | 85.57% |
| TL+DenseNet121 | 93.76% |
| TL+AlexNet | 88.08% |
| TL+VGG11_bn | 88.17% |
| TL+SequezeNet | 92.16% |
| **SDEL detector** | **97.16%** |

In the training phase (namely, the 10-fold stratified cross-validation phase), stochastic gradient descent (SGD) is selected as the optimizer. Learning rate is set to 0.01 and SGD adopts momentum with value of 0.9. Cosine annealing algorithm is adopted to decay the learning rate with the maximization decay moment at the middle time of epochs in each fold and this decay strategy keeps the learning rate varying.

We set 40 epochs in each fold during the training phase. The experimental results of detection accuracy of each base first-tier classifier are shown in **Table 3** (except for the last line). Note that these detection accuracy results will be used in EDTD redistribution weight calculation $p_k/(\sum_{j=0}^n p_j)$ by Eq. (3). After the iterative selecting and testing steps of SDEL approach, ResNet50, VGG16, and GoogLeNet Inception v3 are selected to construct the final SDEL detector. The experiment results indicate:

(1) The detection accuracy of SDEL detector is **97.16%**, which significantly far outperforms the other base first-tier classifiers in **Table 3**.

(2) Grouping the most effective base first-tier classifiers cannot always help achieve a better ensemble learning result. This is indicated by that SDEL detector does not use DenseNet121 which performs best (93.76%) among the eight base first-tier classifiers. This also suggests the effectiveness of using classification distance calculated by Eq. (3) to select the first-tier CNN classifiers. Meanwhile, experimental results validate that different correct classification distributions which complements each other help better enhance ensemble learning results which is consistent with the guidelines of using ensemble learning [36].

(3) More models grouped into ensemble learning may not promise a higher detection accuracy because there are only three first-tier classifiers in SDEL detector.

### 4.3 Evaluation of EDTD Approach

This section first presents explanation heatmaps in order to highlight the importance of metrics defined in Section 3.2.2 and then quantitatively evaluates the quality of explanation methods from the perspectives of Fidelity, Robustness and Expressiveness on heatmaps.

#### 4.3.1 Intuitive demonstration of pixel-level explanation

This section compares the heatmaps generated by a set of explanators, shown in **Fig.2**. The first image is an input image of Adialer.C (one malware family known as a kind of Trojan in Malimg) and the others are heatmaps of input Adialer.C. generated by 13 explanators, including $\nabla o(x)$ [8], $\nabla o(x) \odot x$[8], $(\nabla o(x))^2$ [8], smooth $\nabla o(x)$ [11], guided smooth $\nabla o(x)$ [12], Vanilla $\nabla o(x)$ [9], Occlusion analysis (OA) [14], grad CAM [12], SHAP(positive) [29], SHAP(negative) [29], LIME [19], EDTD-C and EDTD-W. EDTD-W and EDTD-C are two versions of EDTD. EDTD-C merely averages the relevance scores of all base classifiers without using tested detection accuracy as weights in Eq. (12).

$$\widehat{R} = \sum_{0 \leq k < n} R_k / n \qquad (12)$$

EDTD-W uses Eq. (3) to redistribute input relevance differently. Explanator OA uses 90×90 scale occlusion window (red parts represent negative pixels while blue parts represent positive pixels of its image in **Fig.2**). Grad CAM can explain the output of each layer. Here, we only show its heatmap explaining the output of the second-tier classifier in SDEL detector to the input images. LIME plots super-pixels in input images [27]. Both OA and SHAP generates heatmaps by using several colors to denote malicious pixels. But the rest explanators only use red color to represent malicious pixels. In detail, more malicious pixels are represented with lighter and denser red (red can be as light as white), while dark regions are indicative of less relevant pixels.

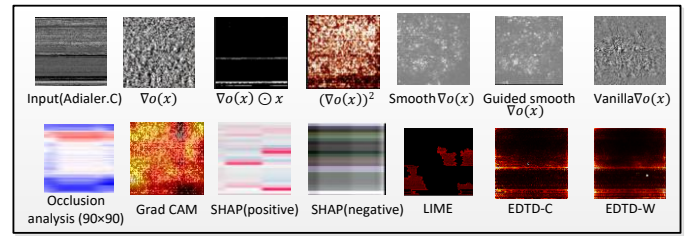

Fig.2 Comparison of heatmaps produced by various image-based explanation methods.

Unlike the MNIST dataset (handwritten numbers) and other normal images that can be understood by humans easily, grayscale malware images are hard to tell which explanation method outperforms in pixel-wise relevance distribution. Nonetheless, we can still roughly infer that EDTD (both EDTD-C and EDTD-W) and $\nabla o(x) \odot x$ seem to more correctly assign most of the relevance to pixels because their heatmap texture is similar to the malware image (the first image) texture. In this respect of observation, although $\nabla o(x) \odot x$ performs great intuitively, it is unable to present the true patterns learned by SDEL detector globally. The other SA-based explanators cannot assign the relevance to malicious pixels. Conversely, SHAP and LIME generate less intuitive heatmaps but it does not mean that they provide distorted answers towards explanation because the principles behind them are solid [19][29].

#### 4.3.2 Quantitative validation results on EDTD

This section uses Fidelity, Robustness and Expressiveness to evaluate each explanator.

Fidelity. To calculate Fidelity, we first plot pixel-flipping curves [8] by feature deduction. **Fig.3** shows the pixel-flipping curves of each baseline explanators and EDTD. In our paper, the pixel-flipping procedure is modified by first removing top10 pixels each time and then ranking them in the descending order of the relevance. This exploiting trick can proactively speed up the flipping. In our modified pixel-flipping procedure, each flip corresponds to setting the pixels in top10 in this time to zero. The range of the heatmap array element value is from 0 to 255.

**Fig.3** is divided into five different regions with different $\tau$, namely, 10, 50, 100, 150, 200 and 250, set by us empirically so as to generate feature sets $\Omega(m, \tau)$ for explanator $m$ by Eq. (4). For each region we remove pixels for the 2, 4, 4, 4, 4, 2 times, respectively, so as to remove target pixels from more relevant to less for totally 20 times on each explanator. In more detail, we remove the targeted pixels in top10 each time and every next time we remove the next top10 pixels. Note if there are no pixels left when pixel removal time in one region is not over, removal stops at this time until the next region begins. With this trick, we then calculate average detection accuracy through all samples by Eq. (5). The average detection accuracy and pixel removal times are respectively set as the ordinate and the abscissa to plot the pixel-flipping curves suited to our case. Regarding the obtained curves of all comparative explanation methods, we find EDTD-W and EDTD-C exhibit respectively the top1 and top2 decreasing speed (sorted in descending order), but SA methods, LIME and SHAP relatively decrease slowly more or less to our intuitive observation. We here intuitively infer that Fidelity of EDTD approach performs the best out of all.

With the plotted curves in **Fig.3**, we use the formula in Eq. (6) to calculate Fidelity of all explanation methods under different $\tau$. The results of **Table 5** show that $FD_m(\tau)$ of both EDTD-W and EDTD-C are the largest, indicating that they outperform other baselines and EDTD-W behaves the best out of all. Note that a higher speed of the curve decreases over pixel flipping times means that the corresponding fidelity score $FD_m(\tau)$ exhibits better. This quantitative validation of Fidelity confirms the inference from the aforementioned intuitive observation. It is worth remarking that the redistribution weights calculated by Eq. (3) strengthen the effect of different base first-tier classifiers on the final output prediction.

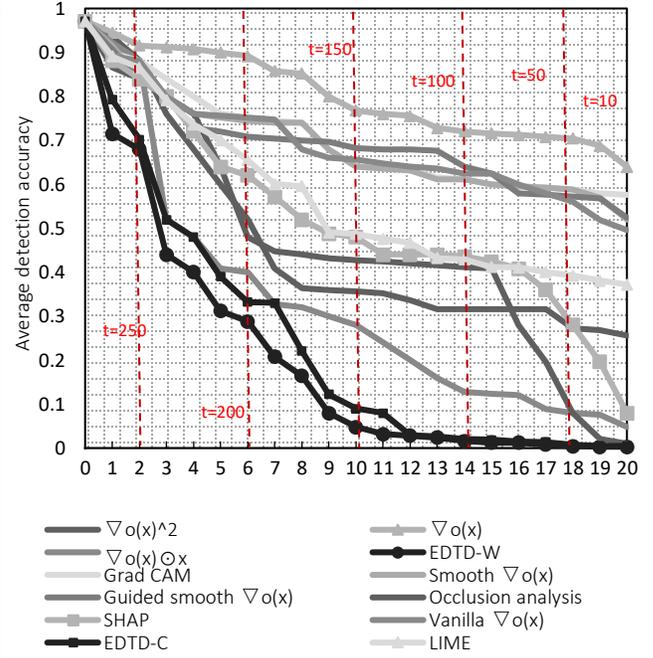

Fig.3 The pixel-flipping curve of average detection accuracy generated by feature removal of top10 in each time of baseline explanators and EDTD (including EDTD-W and EDTD-C) when $\tau$=10, 50, 100, 150, 200, 250.

**Robustness.** To evaluate Robustness, we set $\tau$ in Eq. (4) with 10, 50, 100, 150, 200 and 250 respectively. Theoretically, a more robust explanator exhibits a smaller $RB_m(\tau)$ according to Eq. (8). **Table 4** illustrates the results.

We observe that Robustness of nearly all the explanation increases while $\tau$ increases from 10 to 250 except for the $\nabla o(x) \odot x$. The fewer the number of pixels examined and the higher the relevance of these pixels, the less likely these interpretation results are to be disturbed and the more they can reflect the differences between different classification families. Conversely, when $\tau$ is low, too many noisy features which are unable to explain the prediction affect Robustness. A curious case of $\nabla o(x) \odot x$ in which its Robustness remains unchanged whereas $\tau$ varies indicates that incorporating input image features and gradients may enhance explanation Robustness. However, this trick cannot expose the features correctly learned by detectors but only demonstrating that textures are similar to the original input textures. More generally, Robustness itself cannot suggest the correctness of explanation but the capability of resisting the feature perturbation to explanators. Note that the detailed formulas of Eq. (7,8) suitable for pixel-level explanations are slightly distinctive from those of robustness calculation in [6]. They used semantic feature intersection sets to calculate similarity but we use relevance difference of pixels to measure similarity. But the experimental result trends are same as in [6] with more features of explanations being considered. This same trend indicates that our proposed Robustness formula suitable for pixel-level explanation.

Among all the explanators, $\nabla o(x)$ exhibits the least robustness in all cases. SHAP, LIME and other SA-based methods relatively perform better in all cases. By descending the order of average robustness score of each explanator, the ranking of 12 explanators in term of Robustness is $\nabla o(x) \odot x$ > SHAP > EDTD-W > OA > guided smooth$\nabla o(x)$ > $(\nabla o(x))^2$ > LIME > grad CAM > EDTD-C > vanilla $\nabla o(x)$ > $\nabla o(x)$ > smooth $\nabla o(x)$. This ranking indicates that Robustness can be higher by considering input even in the SA-based methods. However, model-specific explanations like EDTD, guided-type approaches (guided smooth $\nabla o(x)$ and vanilla $\nabla o(x)$) and grad CAM have less robustness because they do not consider element-wise calculations with input images. With this regard, robustness can be stable and high by leveraging the sign and strength of the input. More generally, model-specific explanation methods whose fidelity exhibits relatively good cannot promise better robustness. Conversely, robustness can be high and stable when considering input signs instead of just demonstrating features learned by models.

**Expressiveness**. Based on the above validation results, we comprehensively validate Expressiveness by Eq. (9), as shown in the histograms of **Fig.4**. Each histogram illustrates the variation of Expressiveness of an explanator with the increasing $\tau$. From **Fig.4**, we observe:

(1) SA methods express awful while LIME, SHAP and LRP methods behave relatively better.

(2) Some SA-based methods like $\nabla o(x))^2$, $\nabla o(x) \odot x$ and OA behave suddenly less expressive when $\tau$=250. The possible reason is that they may not be able to explain what SDEL detector truly learns even though their Robustness performs well such as $\nabla o(x) \odot x$. More precisely, these explanators obtain relatively higher expressiveness when $\tau$ is lower than 200 because in these cases feature removal involves with more pixels including those truly critical pixels learned by models but not interpreted by SA-based explanators. Conversely, when $\tau$ is set to 250, the distortion of these explanators is exposed associated with a sudden value drop.

(3) EDTD-W outperforms the other baseline explanators when $\tau$=250, which demonstrates the great expressiveness given by our EDTD-W explanation heatmaps. The corresponding validation indicates that the EDTD-W analysis results yield a reliable outcome by focusing on most malicious pixels exposed by this explanator.

(4) The histogram trend indicates that using fewer features to explain malicious patterns may be more expressive. EDTD behaves rather normally at the beginning when $\tau$ is set with a lower value.

TABLE 4 COMPARISON OF FILDELITY AND ROBUSTNESS OF EXPLANATORS OVER $\tau$

| Explanator | $FD_m(\tau)$ | | | | | | | $RB_m(\tau)$ | | | | | | |
|---|---|---|---|---|---|---|---|---|---|---|---|---|---|---|
| | $\tau$=10 | $\tau$=50 | $\tau$=100 | $\tau$=150 | $\tau$=200 | $\tau$=250 | Avg | $\tau$=10 | $\tau$=50 | $\tau$=100 | $\tau$=150 | $\tau$=200 | $\tau$=250 | Avg |
| $(\nabla o(x))^2$ | 3.58 | 3.89 | 4.68 | 6.16 | 7.53 | 4.50 | **5.06** | 0.6932 | 0.6321 | 0.5928 | 0.5687 | 0.5525 | 0.5414 | **0.5968** |
| $\nabla o(x)$ | 1.66 | 1.49 | 1.80 | 2.04 | 1.33 | 2.78 | **1.85** | 0.8108 | 0.8107 | 0.8105 | 0.8103 | 0.8101 | 0.8092 | **0.8104** |
| $\nabla o(x) \odot x$ | 4.62 | 4.95 | 6.03 | 6.92 | 9.53 | 4.58 | **6.11** | 0.4263 | 0.4263 | 0.4263 | 0.4263 | 0.4263 | 0.4263 | **0.4263** |
| LIME | 3.00 | 3.22 | 3.88 | 4.85 | 5.28 | 5.32 | **4.26** | 0.8079 | 0.7981 | 0.6453 | 0.5986 | 0.4983 | 0.4771 | **0.6376** |
| Grad CAM | 1.98 | 2.13 | 2.57 | 3.32 | 3.66 | 4.58 | **3.04** | 0.8655 | 0.8325 | 0.6988 | 0.6751 | 0.6592 | 0.6487 | **0.7284** |
| Smooth $\nabla o(x)$ | 2.26 | 2.13 | 2.57 | 3.32 | 3.79 | 6.58 | **3.44** | 0.8368 | 0.8209 | 0.8173 | 0.8124 | 0.8097 | 0.8038 | **0.8168** |
| Guided smooth $\nabla o(x)$ | 2.22 | 2.22 | 2.37 | 2.90 | 4.39 | 6.58 | **3.45** | 0.7982 | 0.6421 | 0.5821 | 0.5543 | 0.4923 | 0.4793 | **0.5913** |
| Occlusion analysis | 4.81 | 4.95 | 3.99 | 5.43 | 8.19 | 6.58 | **5.66** | 0.5923 | 0.5432 | 0.5323 | 0.4926 | 0.4783 | 0.4298 | **0.5114** |
| SHAP | 4.46 | 3.84 | 3.83 | 4.92 | 5.86 | 6.58 | **4.92** | 0.4981 | 0.4836 | 0.4368 | 0.4299 | 0.4287 | 0.4283 | **0.4509** |
| Vanilla $\nabla o(x)$ | 2.38 | 2.29 | 2.48 | 3.16 | 3.66 | 4.58 | **3.10** | 0.8039 | 0.8011 | 0.7936 | 0.7921 | 0.7654 | 0.6988 | **0.7758** |
| EDTD-C | 4.83 | 5.36 | 6.80 | 8.82 | 10.66 | 13.48 | **8.33** | 0.8325 | 0.8031 | 0.7744 | 0.7184 | 0.6407 | 0.6162 | **0.7309** |
| EDTD-W | 4.84 | 5.38 | 6.83 | 9.24 | 11.39 | 14.58 | **8.76** | 0.4933 | 0.4683 | 0.4599 | 0.4468 | 0.4381 | 0.4318 | **0.4564** |

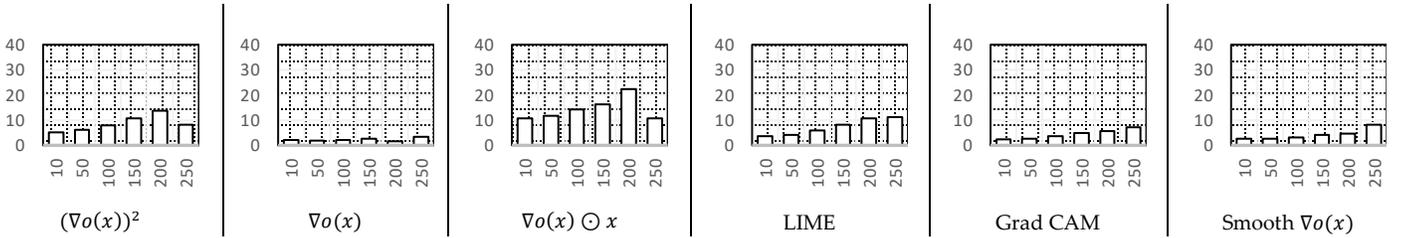

$(\nabla o(x))^2$ | $\nabla o(x)$ | $\nabla o(x) \odot x$ | LIME | Grad CAM | Smooth $\nabla o(x)$

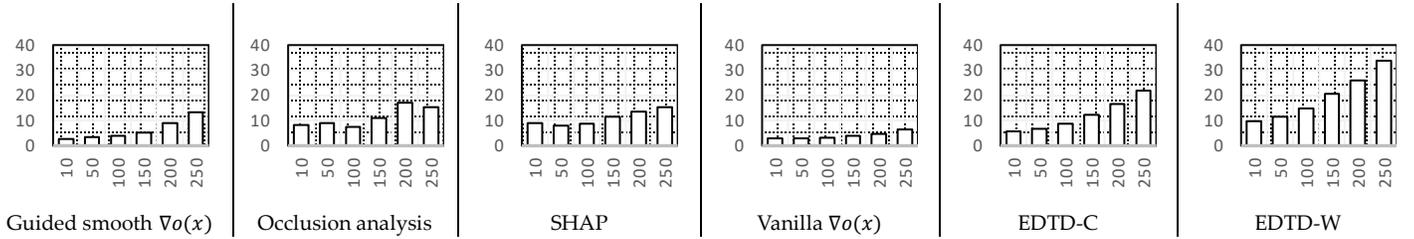

Fig.4 The expressiveness histograms under each explanator (abscissa comes with the value of $\tau$ while ordinate comes with the value of $EP_m(\tau)$ in each subfigure).

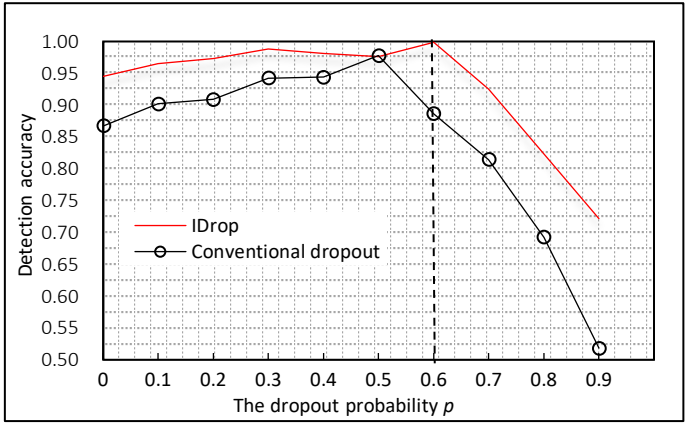

Fig.5 Comparative curves through different dropout methods over dropout probability $p$ (note that $(1-p)$ is multiplied on weights fine-tuned by IDrop and conventional dropout [31] and this trick can rescale the target weights).

(5) Just as mentioned in [5], a larger-scale feature set may yield a better fidelity, but hurts the result interpretability. This assessment of expressiveness quantitatively validates this insight and further helps finding the proper threshold of pixel relevance used in the explanation. Under our experiment configuration, leveraging $(\nabla o(x))^2$ may consider the threshold of 200 instead of 250, leveraging OA may consider 200, LIME may consider 250, and EDTD (both EDTD-W and EDTD-C) may consider 250.

### 4.4 Evaluation of IDrop Approach in Generating IEMD

This section first evaluates the capability of IDrop approach by comparing the detection accuracy of the detectors trained under IDrop approach and a conventional dropout approach (described in Section 2.3), respectively. Then IEMD is compared with the existing image-based malware detectors. Conventional dropout approach (described in Section 2.3) leverages a hyperparameter $p$ to randomly discard the neurons of a layer after the conventional dropout layer. This mechanism requires to pre-set the value of $p$ to conduct 10-fold cross validation and multiply $(1-p)$ to the weights of a layer whose neurons are discarded in the training phase (note training phase refers to 10-fold cross validation) to suit the training scenario [31]. Different from conventional dropout, our IDrop approach does not require selecting $p$ in 10-fold cross validation, and only uses this hyperparameter in the test phase (multiply $(1-p)$ to the weights of a layer whose neurons are discarded based on EDTD approach in the training phase) to make SDEL detector suit the training scenario of discarding neurons based on EDTD approach.

**Fig.5** shows detection accuracy result curves of the detectors trained by IDrop and conventional dropout, respectively. "Conventional dropout" denotes the detection accuracy of the detector which is generated by using conventional dropout to train SDEL detector. "IDrop" denotes IEMD results. The experimental test indicates that there is less variation in the detection accuracy when $0.2 < p < 0.6$ under both IDrop and conventional dropout. When $p$ is adjusted to 0.6, our IEMD detector achieves the best detection accuracy of **0.9987** while the detector achieves 0.9779 when $p$=0.5 in conventional dropout. As the dropout probability $p$ continues increasing, the detection accuracy rate drops drastically because too many neurons are dropped out. We leave the theoretical analysis of the reason for future work. From the results of **Fig.5**, we observe that IDrop outperforms the conventional dropout with random neurons dropped out in [31]. In addition, **Fig.5** validates the effectiveness of using the DTD-based dropout on image datasets, which is questioned in [21].

We now compare IEMD with some previous image-based malware detectors [16][22][23][39] on the Malimg dataset. The performance metric results are summarized in **Table 5** in terms of detection accuracy, predictions, recalls, and F1-scores. The results of [16][22][23][39] are copied from the corresponding papers. Note that their results were obtained by using the whole dataset. Our IEMD's results are on the test set. We observe that IEMD performs better compared to previous works even only on test set.

As mentioned above, there is a conflict between interpretability and detection accuracy. We now specifically depict how IDrop approach mitigates this tradeoff. To validate the above inference, we tend to validate the increasingly remarkable quality of heatmap explanation iteratively generated in our IDrop approach (note that the final heatmaps are stored in **HS** as depicted in **Algorithm 3**) with an increasing detection accuracy.

To implement the aforementioned validation, we manipulate the middle-time EDTD-based heatmaps on the test set when SDEL detector is fine-tuned every 5 epochs. Note that the explanation targets here are the correctly predicted samples from the test set. Heatmaps on the training set are recalculated every 5 epochs during the 10-fold cross validation. **Fig.6** shows that the expressiveness of the recalculated heatmaps generated by EDTD in Step 6 of Algorithm 3 continues to increase when $\tau$=250. The reason of using 250 is that EDTD-W exhibits higher expressiveness shown in **Fig.4** of **Section 4.3.2**. The total epochs are 400 during the 10-fold cross validation, and we recalculate heatmaps by 80 times on the test set. The histogram trend generally increases while jittering up and down during some epochs of heatmap recalculation. Finally, the EDTD-based heatmap explanations can obtain a higher expressiveness of up to 43.12. This overall increasing trend of heatmaps is roughly consistent with that of the detection accuracy during the 10-fold cross validation. The overall trend analyzed by **Fig.6** together with the IEMD detection capability yields mitigation on the tradeoff between interpretability and detection accuracy to some degree.

TABLE 5 COMPARISON OF IMAGE-BASED DETECTION MODELS ON MALIMG

| Ref. | Methods | Detection accuracy | Precision | Recall | F1-score |
|---|---|---|---|---|---|
| [16] | IMCFN (DL) | 0.9882 | 0.9885 | 0.9881 | 0.9875 |
| [22] | DL | 0.9760 | - | 0.8840 | - |
| [23] | k-NN on GIST features | 0.9718 | - | - | - |
| [39] | CNN+SVM on entropy graphs | 0.9972 | - | - | 0.9991 |
| Our paper | **IEMD** | **0.9987** | **0.9980** | **0.9914** | **0.9950** |

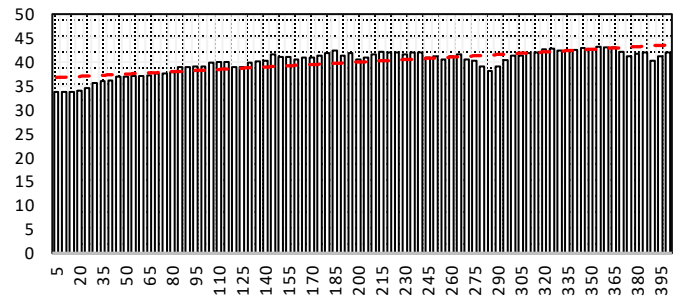

Fig.6 The expressiveness $EP_m(\tau)$ histograms (abscissa comes with epoch number of heatmap recalculation while ordinate comes with the value of $EP_m(\tau)$) on the test set when calulating heatmaps iteratively according to IDrop, and the general trend illustrates that EDTD-based explanation can be increasingly expressive.

## 5 CONCLUSION AND FUTURE WORK

This paper aims to develop an interpretable ensemble learning–based model for image-based malware detection. We first propose an ensemble learning-based detector (SDEL). Then a DTD-based interpretability approach is designed for explain SDEL detector prediction. The explanation results are later used to develop an IDrop approach to improve the detection capability of SDEL detector and then construct an image-based malware detector with high accuracy and high interpretability. We also define formulas for calculating metrics, which are used to evaluate interpretability techniques. Extensive experiment results demonstrate: (1) SDEL detector achieves higher detection accuracy on the Malimg dataset; (2) EDTD approach gives global insights into the ensemble learning model and also outperforms the existing explanators for the Malimg dataset; (3) IDrop can enhance detection performance and help the mitigation of the tradeoff between accuracy and interpretability.

Several future research directions are summarized as follows. (1) In this paper, only 8 base classifiers are investigated to construct the first-tier of SDEL detectors. We plan to explore more base classifiers so as to further improve the detection accuracy. (2) Our experiment results demonstrate that the distance of different base classifiers helps optimize the selection process in SDEL approach. We will explore the theoretical analysis of the reason. (3) The semantic features extracted from pixel-level explanation may discover additional critical malicious behaviors than conventional extraction by static and dynamic analysis. With this regard, we would like to use datasets with original malware files to locate semantic features. Regarding the outcomes of [5, 20, 24], we would also like to compare the located semantic results to theirs in order to find out these additional semantic features extracted from pixel-level explanation.

## REFERENCES

1. F. O. Olowononi, D. B. Rewat, C. Liu: Resilient Machine Learning for Networked Cyber Physical Systems: A Survey for Machine Learning Security to Securing Machine Learning for CPS. IEEE Commun. Surv. Tutorials (2020)..
2. F. Hohman, H. Park, C. Robinson, D. H. (Polo) Chau: Summit: Scaling Deep Learning Interpretability by Visualizing Activation and Attribution Summarizations. IEEE Trans. Vis. Comput. Graph. 26(1): 1096-1106 (2020).
3. R. Guidotti, A. Monreale, S. Ruggieri, F. Turini, F. Giannotti, D. Pedreschi: A Survey of Methods for Explaining Black Box Models. ACM Comput. Surv. 51(5): 93:1-93:42 (2019).
4. D. Vasan, M. Alazab, S. Wassan, B. Safaei, Q. Zheng: Image-Based malware classification using ensemble of CNN architectures (IMCEC). Comput. Secur. 92: 101748 (2020).
5. W. Guo, D. Mu, J. Xu, P. Su, G. Wang, X. Xing: LEMNA: Explaining Deep Learning based Security Applications. CCS 2018: 364-379.
6. M. Fan, W. Wei, X. Xie, Y. Liu, X. Guan, T. Liu: Can We Trust Your Explanations? Sanity Checks for Interpreters in Android Malware Analysis. IEEE Trans. Inf. Forensics Secur. 16: 838-853 (2021).
7. M. T. Ribeiro, S. Singh, C. Guestrin: "Why Should I Trust You?": Explaining the Predictions of Any Classifier. KDD 2016: 1135-1144.
8. J. Kauffmann, K.R. Müller, G. Montavon: Towards explaining anomalies: A deep Taylor decomposition of one-class models. Pattern Recognit. 101: 107198 (2020).
9. D. Erhan, et al. "Visualizing higher-layer features of a deep network." University of Montreal 1341.3 (2009).
10. R. C. Fong, A. Vedaldi: Interpretable Explanations of Black Boxes by Meaningful Perturbation. ICCV 2017: 3449-3457.
11. D. Smilkov, N. Thorat, B. Kim, F. B. Viégas, M. Wattenberg: SmoothGrad: removing noise by adding noise. CoRR abs/1706.03825 (2017).

12. Z. Qi, S. Khorram, F. Li: Visualizing Deep Networks by Optimizing with Integrated Gradients. CVPR Workshops 2019: 1-4.
13. R. R. Selvaraju, M. Cogswell, A. Das, R. Vedantam, D. Parikh, D. Batra: Grad-CAM: Visual Explanations from Deep Networks via Gradient-Based Localization. ICCV 2017: 618-626.
14. M. D. Zeiler, R. Fergus: Visualizing and Understanding Convolutional Networks. CoRR abs/1311.2901 (2013).
15. L. J. Ba, B. J. Frey: Adaptive dropout for training deep neural networks. NIPS 2013: 3084-3092.
16. D. Vasan, M. Alazab, S. Wassan, H. Naeem, B. Safaei, Q. Zheng: IMCFN: Image-based malware classification using fine-tuned convolutional neural network architecture. Comput. Networks 171: 107138 (2020).
17. G. Montavon, S. Lapuschkin, A. Binder, W. Samek, K.R. Müller: Explaining nonlinear classification decisions with deep Taylor decomposition. Pattern Recognit. 65: 211-222 (2017).
18. Z. Xi, G. Panoutsos: Interpretable Machine Learning: Convolutional Neural Networks with RBF Fuzzy Logic Classification Rules. IEEE Conf. on Intelligent Systems 2018: 448-454.
19. S. Chakraborty, R. Tomsett, R. Raghavendra, D. Harborne, M. Alzantot, F. Cerutti, M. B. Srivastava, A. D. Preece, S. Julier, R. M. Rao, T. D. Kelley, D. Braines, M. Sensoy, C. J. Willis, P. Gurram: Interpretability of deep learning models: A survey of results. SmartWorld/SCALCOM/UIC/ATC/CBDCom/IOP/SCI 2017: 1-6.
20. B. Wu, S. Chen, C. Gao, L. Fan, Y. Liu, W. Wen, M. R. Lyu: Why an Android App is Classified as Malware? Towards Malware Classification Interpretation. CoRR abs/2004.11516 (2020).
21. C. Schreckenberger, C. Bartelt, H. Stuckenschmidt: iDropout: Leveraging Deep Taylor Decomposition for the Robustness of Deep Neural Networks. OTM Conferences 2019: 113-126.
22. S. Yajamanam, V. R. S. Selvin, F. D. Troia, M. Stamp: Deep Learning versus Gist Descriptors for Image-based Malware Classification. ICISSP 2018: 553-561.
23. L. Nataraj, S. Karthikeyan, G. Jacob, B. S. Manjunath: Malware images: visualization and automatic classification. VizSEC 2011: 4.
24. M. Q. Li, B. C. M. Fung, P. Charland, S. H. H. Ding: I-MAD: A Novel Interpretable Malware Detector Using Hierarchical Transformer. CoRR abs/1909.06865 (2019).
25. S. Saad, W. Briguglio, H. Elmiligi: The Curious Case of Machine Learning in Malware Detection. ICISSP 2019: 528-535.
26. A. Mills, T. Spyridopoulos, P. Legg: Efficient and Interpretable Real-Time Malware Detection Using Random-Forest. CyberSA 2019: 1-8.
27. L. Chen: Deep Transfer Learning for Static Malware Classification. CoRR abs/1812.07606 (2018).
28. M. Ancona, C. Öztireli, M. H. Gross: Explaining Deep Neural Networks with a Polynomial Time Algorithm for Shapley Values Approximation. CoRR abs/1903.10992 (2019).
29. S. M. Lundberg, S.I. Lee: A Unified Approach to Interpreting Model Predictions. NIPS 2017: 4765-4774.
30. A. Shrikumar, P. Greenside, A. Kundaje: Learning Important Features Through Propagating Activation Differences. ICML 2017: 3145-3153.
31. N. Srivastava, G. E. Hinton, A. Krizhevsky, I. Sutskever, R. Salakhutdinov: Dropout: a simple way to prevent neural networks from overfitting. J. Mach. Learn. Res. 15(1): 1929-1958 (2014).
32. L. Wan, M. D. Zeiler, S. Zhang, Y. L. Cun, R. Fergus: Regularization of Neural Networks using DropConnect. ICML (3) 2013: 1058-1066.
33. S. Wang, X. Wang, P. Zhao, W. Wen, D. R. Kaeli, S. P. Chin, X. Lin: Defensive dropout for hardening deep neural networks under adversarial attacks. ICCAD 2018: 71:1-71:8.
34. D. Bacciu, F. Crecchi: Augmenting Recurrent Neural Networks Resilience by Dropout. IEEE Trans. Neural Networks Learn. Syst. 31(1): 345-351 (2020).
35. R. Keshari, R. Singh, M. Vatsa: Guided Dropout. AAAI 2019: 4065-4072.
36. W. H. Beluch, T. Genewein, A. Nürnberger, J. M. Köhler: The Power of Ensembles for Active Learning in Image Classification. CVPR 2018: 9368-9377.
37. W. Samek, A. Binder, G. Montavon, S. Lapuschkin, K.R. Müller: Evaluating the Visualization of What a Deep Neural Network Has Learned. IEEE Trans. Neural Networks Learn. Syst. 28(11): 2660-2673 (2017).
38. J. Wang, Q. Jing, J. Gao, X. Qiu: SEdroid: A Robust Android Malware Detector using Selective Ensemble Learning. WCNC 2020: 1-5.
39. G. Xiao, J. Li, Y. Chen, K. Li: MalFCS: An effective malware classification framework with automated feature extraction based on deep convolutional neural networks. J. Parallel Distributed Comput. 141: 49-58 (2020).
40. F. N. Iandola, M. W. Moskewicz, K. Ashraf, S. Han, W. J. Dally, K. Keutzer: SqueezeNet: AlexNet-level accuracy with 50x fewer parameters and <1MB model size. CoRR abs/1602.07360 (2016).
41. A. Sengupta, Y. Ye, R. Wang, C. Liu, K. Roy: Going Deeper in Spiking Neural Networks: VGG and Residual Architectures. CoRR abs/1802.02627 (2018).
42. S. M. Lundberg, S.I. Lee: A Unified Approach to Interpreting Model Predictions. NIPS 2017: 4765-4774.
43. S. Yue: Imbalanced Malware Images Classification: a CNN based Approach. CoRR abs/1708.08042 (2017).